\documentclass{article}
\usepackage{spconf,amsmath,amsfonts,graphicx,hyperref}
\usepackage{booktabs}
\usepackage{algorithm}
\usepackage{algpseudocode}
\algrenewcommand\algorithmicrequire{\textbf{Input:}}
\algrenewcommand\algorithmicensure{\textbf{Output:}}
\algrenewcommand\alglinenumber[1]{\fontsize{9}{11}\selectfont #1:}

\usepackage{tikz}
\usetikzlibrary{arrows.meta}

\usepackage{pgfplots}
\usepackage{pgfplotstable}
\usepgfplotslibrary{groupplots}
\pgfplotsset{compat=1.18}
\title{NS-ATTENTION: Newton--Schulz Transformations of Attention Outputs in Vision Transformers}

\name{Xiaohe Jiang \qquad Guoqiang Zhang \qquad Tianjin Huang \qquad Ronghui Mu}
\address{
University of Exeter, Exeter, United Kingdom\\
Emails: \texttt{\{xj272,g.z.zhang,t.huang2,r.mu2\}@exeter.ac.uk}
}

\begin{document}
\ninept
\maketitle

\begingroup
\renewcommand{\thefootnote}{}
\footnotetext{\scriptsize
This work has been submitted to the IEEE for possible publication.
Copyright may be transferred without notice, after which this version
may no longer be accessible.}
\endgroup

\begin{abstract}
Newton--Schulz (NS) iteration has recently been used in the Muon optimizer to transform update matrices during the training of large language models. Motivated by its spectral effect, we investigate applying NS directly to Transformer attention representations. We introduce Newton--Schulz Attention (NS-Attn.), a parameter-free transformation applied to the output of each attention head. Each head output is arranged as a feature-by-token matrix and normalized by its Frobenius norm. We then apply a finite NS polynomial step and restore the original norm. The objective is to reduce spectral concentration and increase effective rank before standard head merging and output projection. Across ViT and Swin on CIFAR-10 and CIFAR-100, NS-Attn. improves final-epoch accuracy in all 12 matched-seed comparisons, with mean gains of 0.25--0.83 percentage points. ViT ablations show higher mean accuracy with one iteration than with two. Spectral analysis further shows reduced leading-eigenvalue concentration and increased effective rank. These gains incur additional inference latency.
\end{abstract}


\vspace{-1mm}
\section{Introduction}
\label{sec:intro}
\vspace{-1mm}

Multi-head attention aggregates value vectors within each head before merging and projecting the outputs~\cite{vaswani2017attention}. Each head's output can be viewed as a feature-by-token matrix, whose squared singular values describe the energy distributed across spectral directions. We investigate whether reshaping this distribution can improve image classification. Our aim is to reduce the dominance of a few directions, producing a flatter normalized eigenvalue spectrum and a higher effective rank.

Newton--Schulz (NS)-style matrix polynomials provide a way to modify singular values using matrix multiplications rather than an explicit singular value decomposition~\cite{jordan2024muon}. Unlike uniform rescaling, these polynomials change the relative magnitudes of different singular values. This motivates their use for redistributing energy in attention representations. We use a finite polynomial transformation rather than requiring exact orthogonalization, and restore the original matrix norm to control the overall output scale. The iteration count therefore becomes a design choice whose effect on classification must be evaluated.

Existing methods apply matrix transformations to different objects. Muon transforms optimizer updates~\cite{jordan2024muon}, while iSQRT-COV and IterNorm use matrix iterations for covariance pooling and feature whitening, respectively~\cite{li2018isqrtcov,huang2019iternorm}. MuonSSM applies NS transformations to low-rank input injections in state-space models~\cite{nguyen2026muonssm}. Within attention, Talking-Heads Attention introduces learned head mixing before and after softmax~\cite{shazeer2020talkingheads}, and spectral conditioning modifies the query, key, and value projection weights~\cite{saratchandran2025spectral}. Orthogonal Self-Attention uses NS iterations to construct a basis from queries and keys for an orthogonal attention operator~\cite{zhang2026orthogonal}. Our approach instead transforms the representations produced by attention-weighted value aggregation.

We introduce \emph{NS Attention}, abbreviated as NS-Attn. To the best of our knowledge, we are the first to apply NS-style iterations directly to per-head Transformer attention outputs, after value aggregation and before head merging. We aim to reduce spectral concentration so that the representation distributes energy more evenly across feature directions. Each head's output is arranged as a feature-by-token matrix. Our main configuration applies one polynomial step, restores the original Frobenius norm, and returns the result to the existing head merging and output projection. The transformation preserves the output shape and introduces no trainable parameters. We refer to the unmodified operation as Standard Attention (Standard Attn.).

We evaluate NS-Attn. using ViT~\cite{dosovitskiy2021vit} and Swin Transformer~\cite{liu2021swin} on CIFAR-10 and CIFAR-100, with three random seeds per setting. Mean final-epoch accuracy improves by 0.25--0.83 percentage points across the four settings, with gains in all 12 matched-seed comparisons. Both backbones show larger gains on CIFAR-100 than on CIFAR-10, suggesting that NS-Attn. may offer greater benefits on more challenging classification tasks. Eigenvalue analysis of trained ViT and Swin Transformer models shows less concentrated output spectra and higher effective rank than Standard Attn., consistent with our design objective. Forward-latency measurements quantify the additional computational cost.

\vspace{-1mm}
\section{Newton--Schulz Transformations of Attention Outputs}
\label{sec:method}
\vspace{-1mm}

\subsection{Attention Output and Feature View}
\label{sec:views}

\begin{figure*}[t]
\centering
\begin{tikzpicture}[
    x=1mm,
    y=1mm,
    font=\fontsize{9}{11}\selectfont,
    nsbox/.style={
        draw=black,
        line width=0.45pt,
        align=center,
        inner sep=2pt
    },
    nsarrow/.style={
        -{Latex[length=1.5mm]},
        line width=0.45pt
    },
    nssum/.style={
        circle,
        draw=black,
        line width=0.45pt,
        minimum size=4mm,
        inner sep=0pt
    }
]
\path[use as bounding box] (0,-7) rectangle (178,64);

% Feature-view matrix.
\node[
    nsbox,
    minimum width=52mm,
    minimum height=9mm
] at (27,52) {
    Attention output\\
    $X\in\mathbb{R}^{B\times H\times N\times D_h}$
};

\node[align=center] at (27,40) {
    NS-Attn.: fix $(b,h)$
};

\node at (33,27) {
    $\begin{bmatrix}
        f_{1,1} & \cdots & f_{1,N} \\
        \vdots & \ddots & \vdots \\
        f_{D_h,1} & \cdots & f_{D_h,N}
    \end{bmatrix}$
};

\node[align=center] at (6,27) {
    Features\\
    $(D_h)$
};

\node at (33,15) {Tokens ($N$)};
\node at (27,-4) {(a) Feature view};

% Transformer block.
\node at (79,60) {Transformer block};
\node (blockinput) at (79,53) {Input};

\node[
    nsbox,
    minimum width=30mm,
    minimum height=9mm
] (attention) at (79,43) {
    Multi-head\\
    attention
};

\node[nssum] (addone) at (79,31) {$+$};

\node[
    nsbox,
    minimum width=30mm,
    minimum height=7mm
] (mlp) at (79,20) {MLP};

\node[nssum] (addtwo) at (79,10) {$+$};
\node (blockoutput) at (79,2) {Output};

\draw[nsarrow] (blockinput.south) -- (attention.north);
\draw[nsarrow] (attention.south) -- (addone.north);
\draw[nsarrow] (addone.south) -- (mlp.north);
\draw[nsarrow] (mlp.south) -- (addtwo.north);
\draw[nsarrow] (addtwo.south) -- (blockoutput.north);

\coordinate (skipone) at (79,50);
\coordinate (skiptwo) at (79,27);
\fill (skipone) circle (0.4mm);
\fill (skiptwo) circle (0.4mm);

\draw[nsarrow]
    (skipone) -- (61,50) -- (61,31) -- (addone.west);
\draw[nsarrow]
    (skiptwo) -- (61,27) -- (61,10) -- (addtwo.west);

% Expansion guides, not computation paths.
\draw[dashed,black!55,line width=0.4pt]
    (attention.north east) -- (112,57);
\draw[dashed,black!55,line width=0.4pt]
    (attention.south east) -- (112,2);

% Attention detail.
\node at (145,60) {Attention detail};

\node[
    nsbox,
    minimum width=50mm,
    minimum height=7mm
] (qkv) at (145,52) {
    $Q$, $K$, $V$ projections
};

\node[
    nsbox,
    minimum width=50mm,
    minimum height=7mm
] (weights) at (145,42) {
    Attention weights $S$
};

\node[
    nsbox,
    minimum width=50mm,
    minimum height=9mm
] (values) at (145,31) {
    Weighted value aggregation\\
    $X=SV$
};

\node[
    nsbox,
    line width=0.9pt,
    fill=black!6,
    minimum width=50mm,
    minimum height=9mm
] (nsinsert) at (145,19) {
    NS-Attn.\\
    Feature view $\to\mathcal{T}\to$ reshape
};

\node[
    nsbox,
    minimum width=50mm,
    minimum height=9mm
] (merge) at (145,7) {
    Head merging\\
    and output projection
};

\draw[nsarrow]
    (qkv.south) -- node[right] {$Q,K$} (weights.north);
\draw[nsarrow] (weights.south) -- (values.north);
\draw[nsarrow] (values.south) -- (nsinsert.north);
\draw[nsarrow] (nsinsert.south) -- (merge.north);

\draw[nsarrow]
    (qkv.east) -- (173,52)
    -- node[right] {$V$} (173,31)
    -- (values.east);

\node at (123,-4) {(b) Transformer context and NS insertion};

\end{tikzpicture}
\vspace{-5mm}
\caption{Feature view and insertion point of NS Attention. (a) For fixed $(b,h)$, the entries $f_{d,n}=X_{b,h,n,d}$ form a feature-by-token matrix. (b) The shaded NS block transforms the attention output before head merging and output projection; Standard Attn. omits this block. Dashed guides expand the attention module, not computation paths. Other unchanged operations are omitted for clarity.}
\label{fig:ns_method}
\end{figure*}
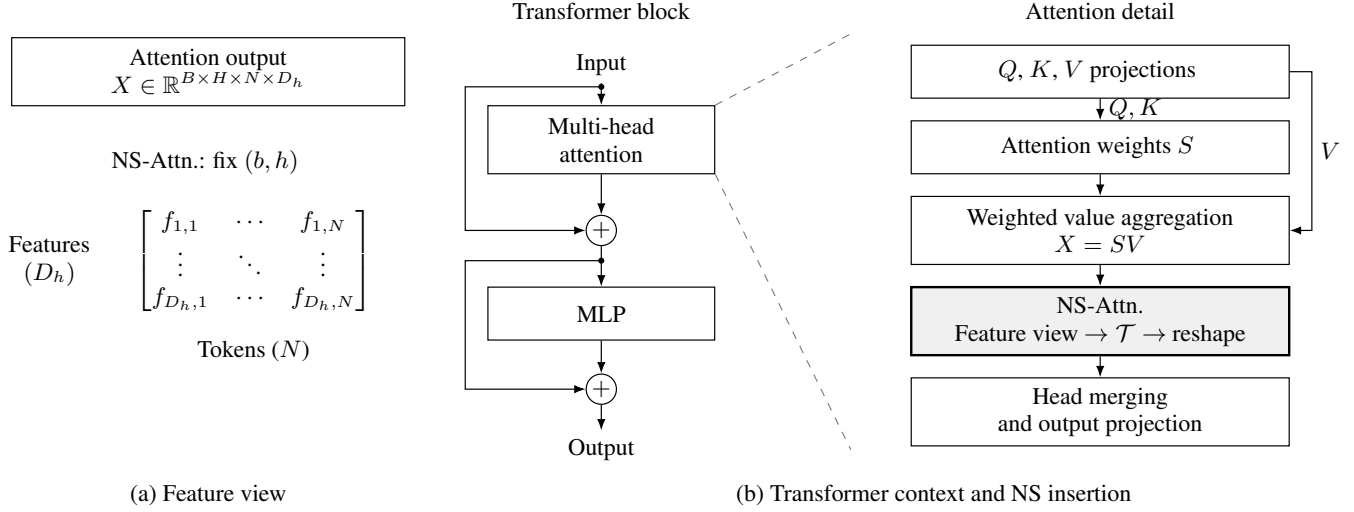

Let $X\in\mathbb{R}^{B\times H\times N\times D_h}$ denote the attention output after weighted value aggregation. Here, $B$ is the batch size, $H$ the number of heads, $N$ the number of tokens, and $D_h$ the features per head. For Swin Transformer, $B$ counts windows across the batch and $N$ is the number of tokens in each window. NS-Attn. transforms $X$ before head merging and output projection, as shown in Fig.~\ref{fig:ns_method}.

For each sample $b$ and head $h$, we transpose the output to form a feature-by-token matrix:
\begin{equation}
M^{(b,h)}
=
\bigl(X_{b,h,:,:}\bigr)^\top
\in\mathbb{R}^{D_h\times N}.
\label{eq:axis_views}
\end{equation}
Each row contains one feature's responses across tokens. The transformation operates on each matrix independently, without mixing outputs from different heads.

\subsection{Newton--Schulz Transformation}
\label{sec:ns_transform}

As we mentioned earlier, NS-Attn. is designed to reduce spectral concentration within each head's feature-by-token matrix. By distributing energy more evenly across spectral directions, we aim to make more effective use of the representation for classification, which may lead to better performance. As will be discussed in Subsection~\ref{sec:spectral},  the trained models indeed exhibit the intended spectral effect.

Next we explain the procedure in detail. For a Feature-view matrix $M$, we first compute its Frobenius norm and normalize it:
\begin{equation}
r=\|M\|_F,
\qquad
\widetilde{M}=\frac{M}{\max(r,\epsilon)},
\label{eq:ns_normalization}
\end{equation}
where $\epsilon=10^{-6}$ is a numerical safeguard against division by very small norms. We form the Gram matrix $A=\widetilde{M}\widetilde{M}^{\top}$ and apply one Muon-inspired polynomial step~\cite{jordan2024muon}:
\begin{equation}
Y=(aI+bA+cA^2)\widetilde{M},
\label{eq:ns_polynomial}
\end{equation}
with $a=3.4445$, $b=-4.7750$, and $c=2.0315$. The entries of $A$ are inner products between feature responses across tokens. The polynomial therefore transforms the features using relationships measured within the same head.

We then normalize $Y$ and restore the original norm:
\begin{equation}
\mathcal{T}(M)
=
r\,\frac{Y}{\max(\|Y\|_F,\epsilon)}.
\label{eq:ns_rescaling}
\end{equation}
When the stabilizing clamps are inactive, $\|\mathcal{T}(M)\|_F=\|M\|_F$. The operation changes the singular-value distribution while preserving the matrix norm. We use NS to denote this finite Muon-inspired transformation, rather than exact orthogonalization.

The result is transposed back to the original layout:
\begin{equation}
\widehat{X}_{b,h,:,:}
=
\left[\mathcal{T}\left(M^{(b,h)}\right)\right]^\top.
\label{eq:feature_ns}
\end{equation}
The tensor $\widehat{X}$ has the same shape as $X$ and passes through the existing head merging and output projection. Standard Attn. passes $X$ through these operations unchanged. No trainable parameters are added.

Algorithm~\ref{alg:ns_feature} describes NS-Attn. with $K$ polynomial iterations. We use $K=1$ in the main method and compare it with $K=2$ in the ablation study. Normalization and norm restoration are each performed once, before and after the iteration loop, respectively. 

We now briefly discuss the computational complexity of the NS iteration. Since each Gram matrix has shape $D_h\times D_h$, this leads to a total arithmetic cost of $O\!\left(KBH(D_h^2N+D_h^3)\right)$ for the Gram matrix and polynomial operations. It is worth noting that operations across samples and heads can be evaluated in parallel using batched matrix products. 

\begin{algorithm}[t]
\caption{NS-Attn. with $K$ iterations}
\label{alg:ns_feature}
\fontsize{9}{11}\selectfont
\begin{algorithmic}[1]
\Require Attention output $X\in\mathbb{R}^{B\times H\times N\times D_h}$
\Require Iteration count $K\geq1$ (default: $1$)
\Ensure Transformed output $\widehat{X}$ with the same shape as $X$
\Statex $a=3.4445,\ b=-4.7750,\ c=2.0315$
\Statex $\epsilon=10^{-6}$
\ForAll{sample indices $s$ and head indices $h$}
    \State $M\gets (X_{s,h,:,:})^\top$
    \State $r\gets\|M\|_F$
    \State $Z\gets M/\max(r,\epsilon)$
    \For{$k=1,\ldots,K$}
        \State $A\gets ZZ^\top$
        \State $Z\gets aZ+(bA+cA^2)Z$
    \EndFor
    \State $\widehat{M}\gets rZ/\max(\|Z\|_F,\epsilon)$
    \State $\widehat{X}_{s,h,:,:}\gets\widehat{M}^{\top}$
\EndFor
\State \Return $\widehat{X}$
\end{algorithmic}
\end{algorithm}

\vspace{-1mm}
\section{Experiments and Analysis}
\label{sec:results}
\vspace{-1mm}

\subsection{Experimental Setup}
\label{sec:setup}

\textbf{Datasets and architectures.} We evaluate Standard Attn. and NS-Attn. on CIFAR-10 and CIFAR-100~\cite{krizhevsky2009cifar} using ViT~\cite{dosovitskiy2021vit} and Swin Transformer~\cite{liu2021swin}.\footnote{ViT and Swin code: \url{https://github.com/aanna0701/SPT_LSA_ViT}; SPT and LSA disabled.} ViT uses $4\times4$ patches, an embedding dimension of 192, and nine blocks. Each block has 12 attention heads with $D_h=16$ and an MLP expansion ratio of 2. All nine blocks use a fixed stochastic-depth rate of 0.1. Swin uses $2\times2$ patches and three stages with depths $[2,6,4]$. Their embedding dimensions are $[96,192,384]$, with $[3,6,12]$ attention heads, respectively. The head dimension is 32, and each attention window contains $4\times4$ tokens. The stochastic-depth rate increases linearly from 0 to 0.1 across all 12 blocks. NS-Attn. is applied in every attention block of both backbones.

\noindent \textbf{Training setup.} All models are trained for 100 epochs with AdamW~\cite{loshchilov2019adamw} and a batch size of 128. The peak learning rate is $10^{-3}$ and weight decay is $0.05$. A 10-epoch warm-up is followed by cosine learning-rate decay. All configurations use random cropping, horizontal flipping, CIFAR AutoAugment~\cite{cubuk2019autoaugment}, random erasing~\cite{zhong2020randomerasing}, Mixup~\cite{zhang2018mixup}, CutMix~\cite{yun2019cutmix}, and repeated augmentation~\cite{hoffer2020batch}. We also use label smoothing~\cite{szegedy2016inception} and dataset-specific input normalization.

\noindent \textbf{Evaluation protocol.} We report final-epoch Top-1 accuracy on the official test set, averaged over seeds 0, 1, and 3, together with the sample standard deviation. Each comparison pairs models trained with the same seed. Accuracy gains are computed from unrounded means and reported in percentage points relative to Standard Attn.

\subsection{Classification Performance}
\label{sec:feature_validation}

Table~\ref{tab:main_results} compares Standard Attn. and NS-Attn. across the two backbones and datasets. NS-Attn. improves mean final-epoch accuracy in all four settings. Accuracy also improves in all 12 matched seed comparisons.

\begin{table}[h]
\centering
\setlength{\tabcolsep}{3pt}
\caption{Final-epoch Top-1 accuracy (\%), reported as mean $\pm$ sample SD across seeds 0, 1, and 3. Gains are in percentage points and are computed before rounding.}
\label{tab:main_results}
\vspace{3pt}
\begin{tabular*}{\columnwidth}{@{\extracolsep{\fill}}llccc@{}}
\toprule
Backbone & Dataset & Standard & NS-Attn. (\textbf{Ours}) & Gain \\
         &         & Attn.    &          & (pp) \\
\midrule
ViT  & CIFAR-10  & $93.50 \pm 0.11$ & $\mathbf{93.88} \pm 0.20$ & $+0.38$ \\
ViT  & CIFAR-100 & $72.43 \pm 0.21$ & $\mathbf{73.25} \pm 0.49$ & $+0.82$ \\
Swin & CIFAR-10  & $95.10 \pm 0.10$ & $\mathbf{95.35} \pm 0.06$ & $+0.25$ \\
Swin & CIFAR-100 & $77.37 \pm 0.11$ & $\mathbf{78.19} \pm 0.30$ & $+0.83$ \\
\bottomrule
\end{tabular*}
\end{table}

The gains occur with both global attention in ViT and shifted-window attention in Swin. Both backbones show larger absolute improvements on CIFAR-100 than on CIFAR-10, suggesting that NS-Attn. may offer greater benefits on more challenging classification tasks.

\subsection{Ablation Study}
\label{sec:ablation}

\noindent\textbf{Impact of NS iteration $K$}: Muon uses five NS iterations to approximately orthogonalize optimizer updates~\cite{jordan2024muon}. Here, the transformation acts on attention outputs instead, so we examine whether an additional iteration improves classification accuracy. We compare Standard Attn. with one and two NS iterations on ViT using three random seeds for each dataset. One iteration is our main configuration.

All three configurations use matching seeds and identical training settings.

\begin{table}[h]
\centering
\setlength{\tabcolsep}{3pt}
\caption{Iteration-count ablation on ViT. Final-epoch Top-1 accuracy (\%) is reported as mean $\pm$ sample SD across three random seeds. The iteration counts refer to NS-Attn.}
\label{tab:iteration_ablation}
\vspace{3pt}
\begin{tabular*}{\columnwidth}{@{\extracolsep{\fill}}lccc@{}}
\toprule
Dataset & Standard & 1 iteration & 2 iterations \\
        & Attn.    &             &              \\
\midrule
CIFAR-10  & $93.50 \pm 0.11$ & $\mathbf{93.88} \pm 0.20$ & $93.49 \pm 0.08$ \\
CIFAR-100 & $72.43 \pm 0.21$ & $\mathbf{73.25} \pm 0.49$ & $73.05 \pm 0.73$ \\
\bottomrule
\end{tabular*}
\end{table}

Table~\ref{tab:iteration_ablation} shows that a second iteration lowers mean accuracy by 0.39 percentage points on CIFAR-10 and 0.19 points on CIFAR-100, computed before rounding. On CIFAR-10, accuracy decreases in all three matched seed comparisons and returns close to Standard Attn. On CIFAR-100, two iterations retain a 0.62-point mean gain over Standard Attn., but outperform one iteration in only one of the three seeds. The sample standard deviation also increases from 0.49 to 0.73 points on CIFAR-100. A possible explanation is that a second polynomial update changes the relative weighting of spectral directions in a way that is less favorable for classification. Overall, one iteration achieves higher mean accuracy on both datasets with fewer polynomial updates, supporting its use as our main configuration.

\noindent\textbf{Impact of hyper-parameter $\epsilon$}: We also examine sensitivity to $\epsilon$, which safeguards the two Frobenius-norm divisions. We checked six values from $10^{-8}$ to $10^{-2}$ using the final seed-0 ViT checkpoints on both datasets and 160 test images per dataset. Neither clamp was activated for any candidate threshold. Direct CPU FP32 comparisons at $10^{-4}$, $10^{-6}$, and $10^{-8}$ produced element-wise identical transformed outputs and logits. We also monitored the first five training epochs of a CIFAR-100 seed-0 run at the default $\epsilon=10^{-6}$. Neither norm fell below any candidate threshold during the 5,855 training steps. Thus, the transformation was insensitive to the tested thresholds in the audited final-checkpoint evaluations and early-training computations.

\subsection{Spectral Analysis}
\label{sec:spectral}

We perform spectral analysis using the final checkpoints of Standard Attn. and NS-Attn. for both ViT and Swin Transformer on CIFAR-10 and CIFAR-100. Each checkpoint receives the same first 160 test images in five batches of 32, without shuffling. We use evaluation mode and \texttt{torch.inference\_mode()}. Outputs are collected immediately before head merging and output projection. For NS-Attn., these are the transformed outputs.

% Requires tikz, pgfplots, pgfplotstable, and \usepgfplotslibrary{groupplots}.
% Supplied mean spectra are plotted directly; no renormalization or reaggregation.
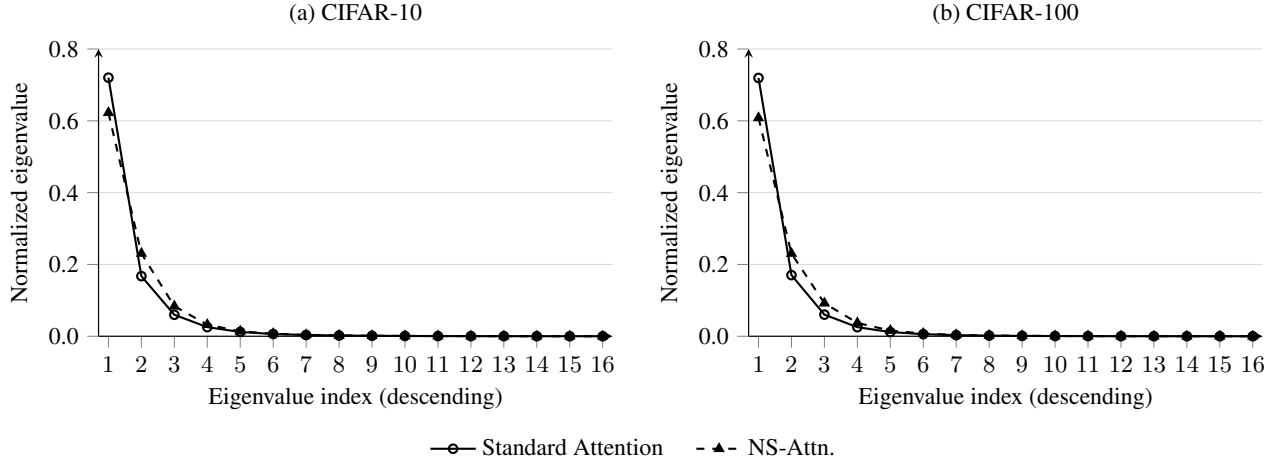
\begin{figure*}[t]
\centering
\pgfplotstableread[col sep=space,row sep=\\]{
index Std10 NS10 Std100 NS100\\
1 0.720222670963 0.622480252354 0.719409831838 0.608091751812\\
2 0.167393779643 0.230614770028 0.170295449938 0.230363508974\\
3 0.059871275200 0.083659530875 0.060256949369 0.091972057042\\
4 0.025450757757 0.032310385325 0.025306747457 0.037246459980\\
5 0.011955253298 0.013783773693 0.011716559498 0.015911113144\\
6 0.006232236820 0.006992483985 0.005830769916 0.007426484546\\
7 0.003504371833 0.003936097169 0.003096167303 0.003868778407\\
8 0.002063658543 0.002337293407 0.001718662587 0.002132771585\\
9 0.001263843645 0.001442779943 0.000988616535 0.001226079606\\
10 0.000789527640 0.000908516589 0.000576843325 0.000720036483\\
11 0.000502139520 0.000588964227 0.000342593937 0.000432553416\\
12 0.000318775447 0.000383669627 0.000205451238 0.000263056126\\
13 0.000200672595 0.000249897585 0.000122590705 0.000160584108\\
14 0.000123793849 0.000159839104 0.000071920784 0.000097254797\\
15 0.000071864321 0.000097846704 0.000040639574 0.000057157718\\
16 0.000035378926 0.000053899383 0.000020205997 0.000030352255\\
}\nsspectrumtable

\begin{tikzpicture}
\begin{groupplot}[
    group style={group size=2 by 1,horizontal sep=18mm},
    width=68mm,
    height=38mm,
    scale only axis,
    xmin=0.7,
    xmax=16.3,
    ymin=0,
    ymax=0.8,
    xtick={1,2,3,4,5,6,7,8,9,10,11,12,13,14,15,16},
    ytick={0,0.2,0.4,0.6,0.8},
    yticklabels={0.0,0.2,0.4,0.6,0.8},
    xlabel={Eigenvalue index (descending)},
    ylabel={Normalized eigenvalue},
    axis lines=left,
    tick align=outside,
    ymajorgrids=true,
    grid style={black!15,line width=0.3pt},
    /tikz/line width=0.45pt,
    /tikz/font={\fontsize{9}{11}\selectfont},
    tick label style={font=\fontsize{9}{11}\selectfont},
    label style={font=\fontsize{9}{11}\selectfont},
    title style={font=\fontsize{9}{11}\selectfont},
    legend style={
        draw=none,
        font=\fontsize{9}{11}\selectfont,
        /tikz/every even column/.append style={column sep=10pt}
    },
    legend cell align=left,
    legend columns=2,
    clip=false
]
\nextgroupplot[
    title={(a) CIFAR-10},
    legend to name=ns-spectrum-legend
]
\addplot[
    black,solid,line width=0.8pt,
    mark=o,mark size=1.6pt,
    mark options={solid,fill=white}
] table[x=index,y=Std10] {\nsspectrumtable};
\addlegendentry{Standard Attention}

\addplot[
    black,dashed,line width=0.8pt,
    mark=triangle*,mark size=1.7pt,
    mark options={solid,fill=black}
] table[x=index,y=NS10] {\nsspectrumtable};
\addlegendentry{NS-Attn.}

\nextgroupplot[title={(b) CIFAR-100}]
\addplot[
    black,solid,line width=0.8pt,
    mark=o,mark size=1.6pt,
    mark options={solid,fill=white}
] table[x=index,y=Std100] {\nsspectrumtable};

\addplot[
    black,dashed,line width=0.8pt,
    mark=triangle*,mark size=1.7pt,
    mark options={solid,fill=black}
] table[x=index,y=NS100] {\nsspectrumtable};
\end{groupplot}
\end{tikzpicture}

\par\smallskip
\pgfplotslegendfromname{ns-spectrum-legend}
\vspace{-2mm}
\caption{Mean normalized eigenvalues of Feature-view Gram matrices from final ViT attention outputs. Each spectrum is sorted in descending order and normalized before averaging over seeds 0, 1, and 3, nine blocks, 160 test images, and 12 heads (51,840 spectra per curve).}
\label{fig:feature_spectrum}
\vspace{-3mm}
\end{figure*}

For ViT, each sample and head yields a feature-by-token matrix $M\in\mathbb{R}^{16\times65}$. We analyze its Gram matrix:
\begin{equation}
G=MM^\top\in\mathbb{R}^{16\times16}.
\label{eq:spectral_gram}
\end{equation}
Its eigenvalues are the squared singular values of $M$. We sort and normalize them:
\begin{equation}
p_i=\frac{\lambda_i}{\sum_{j=1}^{16}\lambda_j},
\qquad
\lambda_1\geq\cdots\geq\lambda_{16}.
\label{eq:normalized_spectrum}
\end{equation}
The leading-eigenvalue fraction $p_1$ measures the energy share of the dominant direction. Following Roy and Vetterli~\cite{roy2007effective}, we compute the entropy effective rank of the Gram matrix $G$ as
\begin{equation}
r_{\mathrm{eff}}
=
\exp\left(-\sum_{i=1}^{16}p_i\log p_i\right),
\label{eq:effective_rank}
\end{equation}
where $\log$ denotes the natural logarithm and $0\log0=0$. Higher effective rank indicates a more even energy distribution. For Swin, each window and head yields a $32\times16$ matrix; the same definitions apply to all 32 eigenvalues of its $32\times32$ Gram matrix.

Across three seeds, each dataset and attention variant provides 51,840 spectra for ViT and 138,240 for Swin. Each spectrum is normalized individually, and both metrics are computed per matrix before aggregation. ViT metrics are averaged across matrices. Swin metrics are summarized by block within each seed and then averaged across seeds. Figure~\ref{fig:feature_spectrum} plots the mean normalized ViT spectra; effective rank is not computed from the mean spectrum.

For ViT, Fig.~\ref{fig:feature_spectrum} shows a smaller leading-eigenvalue fraction and greater energy shares in subsequent directions with NS-Attn. On CIFAR-10, mean $p_1$ decreases from 0.72 to 0.62, while mean $r_{\mathrm{eff}}$ increases from 2.53 to 2.91. On CIFAR-100, the corresponding changes are 0.72 to 0.61 and 2.49 to 3.00.

Swin Transformer shows a similar pattern. On CIFAR-10, mean $p_1$ decreases from 0.78 to 0.64, while mean $r_{\mathrm{eff}}$ increases from 2.21 to 3.04. On CIFAR-100, the corresponding changes are 0.79 to 0.61 and 2.19 to 3.13.

\subsection{Computational Cost}
\label{sec:cost}

We benchmark GPU forward latency on an NVIDIA GeForce RTX 4060 Laptop GPU using PyTorch 2.8.0 (CUDA 12.8), FP32 eager execution, evaluation mode, and \texttt{torch.inference\_mode()}. Inputs are the first 128 normalized CIFAR-100 test images ($32\times32$), already resident on the GPU. Within each backbone, both configurations load identical parameters and buffers from the final seed-0 NS-Attn. checkpoint, isolating implementation overhead rather than comparing separately trained models.

Three rounds each use 50 warm-up and 200 timed forwards, alternating configuration order between rounds. Synchronized CUDA events measure model execution only, excluding preprocessing, data loading, and host-to-device transfer.

\begin{table}[t]
\centering
\setlength{\tabcolsep}{3pt}
\caption{GPU forward latency (ms/batch; batch size 128). Means cover 600 forwards; $\pm$ denotes the sample SD across three round means. Overhead is relative to Standard Attn.}
\label{tab:inference_cost}
\vspace{3pt}
\begin{tabular*}{\columnwidth}{@{\extracolsep{\fill}}lccc@{}}
\toprule
Backbone & Standard & NS-Attn. (\textbf{Ours}) & Overhead \\
         & Attn.    &          &          \\
\midrule
ViT  & $23.55 \pm 0.45$ & $33.20 \pm 0.74$ & $+41.0\%$ \\
Swin & $23.81 \pm 0.04$ & $48.51 \pm 0.36$ & $+103.7\%$ \\
\bottomrule
\end{tabular*}
\end{table}

Table~\ref{tab:inference_cost} shows that NS-Attn. increases GPU forward latency by 41.0\% for ViT and 103.7\% for Swin in the measured configuration. Its additional matrix operations therefore introduce a clear accuracy--latency trade-off despite adding no trainable parameters, with a larger relative overhead for Swin.

\vspace{-1mm}
\section{Discussion and Limitations}
\label{sec:discussion}
\vspace{-1mm}

NS-Attn. targets the distribution of energy within each head's output. Restoring the input Frobenius norm preserves the total spectral energy when the numerical clamps are inactive, while the polynomial changes its distribution across directions. In the trained ViT and Swin Transformer models, the smaller leading-eigenvalue fraction and higher effective rank are consistent with the intended reduction in spectral concentration. These measurements describe a representation effect that accompanies the classification gains; they do not establish that spectral flattening causes those gains.

The classification improvements and aggregate eigenvalue trends extend to both ViT and Swin Transformer. The additional forward latency also shows that adding no trainable parameters does not make the transformation computationally free. Its practical value therefore depends on the accuracy--latency requirements of the application. Evaluation on larger-scale datasets and under equal compute budgets would help establish the scope of the observed benefits.

\vspace{-1mm}
\section{Conclusion}
\label{sec:conclusion}
\vspace{-1mm}

We introduced NS-Attn., which applies a finite Newton--Schulz-style transformation to each head's attention output without adding trainable parameters. One iteration improves mean final-epoch accuracy across the evaluated ViT and Swin settings. Eigenvalue analysis of trained ViT and Swin Transformer models shows less concentrated output spectra and higher effective rank.

\bibliographystyle{IEEEbib}
\bibliography{references}

\end{document}